\documentclass[10pt,twocolumn,twoside]{IEEEtran}
\pdfoutput=1

\usepackage{times}
\usepackage{amssymb}
\usepackage{graphicx}
\usepackage{algorithmic}
\usepackage{float}
\usepackage{multirow}
\usepackage{array}
\usepackage{balance}
\usepackage[pagebackref=false,breaklinks=true,colorlinks,bookmarks=false,urlcolor=blue,linkcolor=blue,citecolor=blue,pdftitle={A Low-Complexity Algorithm for Static Background Estimation from Cluttered Image Sequences in Surveillance Contexts},pdfauthor={Vikas Reddy, Conrad Sanderson, Brian C. Lovell}]{hyperref}

\begin{document}

\pagestyle{empty}

\title
  {
  \bf\LARGE
  A Low-Complexity Algorithm for Static Background Estimation\\
  from Cluttered Image Sequences in Surveillance Contexts
  }

\author
  {
  ~\\
  {\Large  Vikas Reddy, Conrad Sanderson, Brian C. Lovell}% <-this % stops a space
  \thanks
    {
    \hrule

    \hspace{1ex}

    {\bf Published in:} \mbox{EURASIP Journal on Image and Video Processing,~2011.}
    \href{http://dx.doi.org/10.1155/2011/164956}{\bf http://dx.doi.org/10.1155/2011/164956}\newline
    \mbox{C++ source code available at} \href{http://arma.sourceforge.net/background\_est/}{\bf http://arma.sourceforge.net/background\_est/}
    }

  ~\\

  NICTA, PO Box 6020, St Lucia, QLD 4067, Australia\\
  University of Queensland, School of ITEE, QLD 4072, Australia\\
  }

\maketitle
\thispagestyle{empty}

\begin{abstract}

For the purposes of foreground estimation,
the true background model is unavailable in many practical circumstances and needs to be estimated from cluttered image sequences.
We propose a sequential technique
for static background estimation in such conditions, with low computational and memory requirements.
Image sequences are analysed on a block-by-block basis.
For each block location a representative set is maintained which contains distinct blocks obtained along its temporal line.
The background estimation is carried out in a Markov Random Field framework,
where the optimal labelling solution is computed using iterated conditional modes.
The clique potentials are computed based on the combined frequency response of the candidate block and its neighbourhood.
It is assumed that the most appropriate block results in the smoothest response,
indirectly enforcing the spatial continuity of structures within a scene.
Experiments on real-life surveillance videos demonstrate that the proposed method obtains considerably better background estimates
(both qualitatively and quantitatively)
than median filtering and the recently proposed ``intervals of stable intensity'' method.
Further experiments on the Wallflower dataset suggest that the combination of the proposed method
with a foreground segmentation algorithm results in improved foreground segmentation.

\end{abstract}

\vspace{-2ex}
\section{Introduction}

Intelligent surveillance systems can be used effectively for monitoring critical infrastructure
such as banks, airports and railway stations~\cite{wolf2002sce}.
Some of the key tasks of these systems are real-time segmentation,
tracking and analysis of foreground objects of interest~\cite{Collins_2000_3325,Sanderson_ICB_2009}.
Many approaches for detecting and tracking objects are based on background subtraction techniques,
where each frame is compared against a background model for foreground object detection.

The majority of background subtraction methods
adaptively model and update the background for every new input frame.
Surveys on this class of algorithms are found in~\cite{cheung2004rtb,piccardi2004bst}.
However, most methods presume the training image sequence used to model the background is free from foreground objects
~\cite{1597122,vargas-enhanced,matsuyama2006background}.
This assumption is often not true in the case of uncontrolled environments such as train stations and airports,
where directly obtaining a clear background is almost impossible.
Furthermore, in certain situations a strong illumination change can render the existing background model ineffective,
thereby forcing us to compute a new background model.
In such circumstances,
it becomes inevitable to estimate the background using cluttered sequences (i.e.~where parts of the background are occluded).
A good background estimate will complement the succeeding background subtraction process, which can result
in improved detection of foreground objects.

\begin{figure}[!t]
  \centering

  \begin{minipage}{0.95\columnwidth}
    \begin{minipage}{.49\columnwidth}
      \includegraphics[width=1\columnwidth]{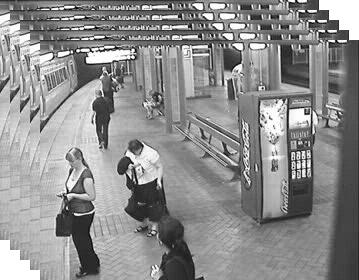}
    \end{minipage}
    \hfill
    \begin{minipage}{.49\columnwidth}
      \includegraphics[width=\columnwidth]{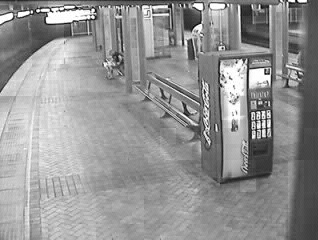}
    \end{minipage}
  \end{minipage}

  \centerline{~}

  \begin{minipage}{0.95\columnwidth}
    \begin{minipage}{.49\columnwidth}
      \centerline{\bf(i)}
    \end{minipage}
    \hfill
    \begin{minipage}{0.49\columnwidth}
      \centerline{\bf(ii)}
    \end{minipage}
  \end{minipage}
  \vspace{-1ex}
  \caption
    {
    \small
    Typical example of estimating the background
    from an cluttered image sequence:
    {\bf (i)}~input frames cluttered with foreground objects, where only parts of the background are visible;
    {\bf (ii)}~estimated background.
    }
  \label{fig:BGInit_EX}
\vspace{-3ex}
\end{figure}

The problem can be paraphrased as follows:
given a short image sequence captured from a stationary camera
in which the background is occluded by foreground objects in every frame of the sequence for most
of the time,
the aim is to estimate its background, as illustrated in Figure~\ref{fig:BGInit_EX}.
This problem is also known in the literature as background initialisation or bootstrapping~\cite{toyama1999wpa}. Background
estimation is related to, but distinct from, background modelling. Owing to the complex nature of the problem,
we confine our estimation strategy to static backgrounds (e.g. no waving trees),
which is quite common in urban surveillance environments
such as banks, shopping malls, airports and train stations.

Existing background estimation techniques, such as simple median filtering,
typically require the storage of all the input frames in memory before estimating the background.
This increases memory requirements immensely.
In this paper we propose a robust background estimation algorithm in a Markov Random Field
(MRF) framework.
It operates on the input frames sequentially, avoiding the need to store all the frames.
It is also computationally less intensive, enabling the system to achieve real-time performance ---
this aspect is critical in video surveillance applications.
This paper is a thoroughly revised and extended version of our previous work~\cite{vreddysbe2009}.

We continue as follows.
Section~\ref{sec:Previous work} gives an overview of existing methods for background estimation.
Section~\ref{sec:Proposed Algorithm} describes the proposed algorithm in detail.
Results from experiments on real-life surveillance videos are given in Section~\ref{sec:Experimental Results},
followed by the main findings in Section~\ref{sec:Conclusion}.

\section{Previous Work}
\label{sec:Previous work}

Existing methods to address the cluttered background estimation problem can be broadly classified into three categories:
{\bf (i)}~pixel-level processing,
{\bf (ii)}~region-level processing,
{\bf(iii)}~a hybrid of the first two.
It must be noted that all methods assume the background to be static.
The three categories are overviewed in the sections below.

\subsection{Pixel-level Processing}
\label{subsec_pixel-level processing}
In the first category the simplest techniques are based on applying a median filter
on pixels at each location across all the frames.
Lo and Velastin~\cite{lo2001acd} apply this method to obtain reference background
for detecting congestion on underground train platforms.
However, its limitation is that the background is estimated correctly
only if it is exposed for more than 50\% of the time.
Long and Yang~\cite{long1990sbg} propose an algorithm that finds pixel intervals of stable intensity in the image sequence,
then heuristically chooses the value of the longest stable interval to most likely represent the background.
Bevilacqua~\cite{bevilacqua2002nbi} applies Bayes' theorem in his proposed approach.
For every pixel it estimates the intensity value to which that pixel has the maximum posterior probability.

Wang and Suter~\cite{wang2006nrs} employ a two-staged approach. The first stage is similar to that of~\cite{long1990sbg},
followed by choosing background pixel values whose interval maximises an objective function.
It is defined as ${N_{l_k}}/{S_{l_k}}$  where  $N_{l_k}$ and $S_{l_k}$
are the length and standard variance of the $k$-th  interval of pixel sequence~$l$.
The method proposed by Kim et al.~\cite{kim2005rtf} quantises the temporal values of each
pixel into distinct bins called codewords.
For each codeword, it keeps a record of the maximum time interval during which it has not recurred.
If this time period is greater than $N/2$, where $N$ is the total number of frames in the sequence,
the corresponding codeword is discarded as foreground pixel.
The system recently proposed by Chiu et al.~\cite{chiu2010robust} estimates the background
and utilises it for object segmentation. Pixels obtained from each location 
along its time axis are clustered based on a threshold. The pixel corresponding to 
the cluster having the maximum probability and greater than a time-varying
threshold is extracted as background pixel. 

All these pixel based techniques can perform well when the foreground objects are moving,
but are likely to fail when the time interval of exposure of the background is less than that of the foreground.

\subsection{Region-level Processing}
\label{subsec_region-level processing}
In the second category, the method proposed by
Farin et al.~\cite{farin2003phw} performs a rough segmentation of input frames into foreground and background regions.
To achieve this, each frame is divided into blocks, the temporal sum of absolute differences (SAD)
of the co-located blocks is calculated, and a block similarity matrix is formed.
The matrix elements that correspond to small SAD values are considered as stationary elements
and high SAD values correspond to non-stationary elements.
A~median filter is applied only on the blocks classified as background.
The algorithm works well in most scenarios, however,
the spatial correlation of a given block with its neighbouring blocks already filled by background
is not exploited, which can result in estimation errors
if the objects  are quasi-stationary for extended periods.

In  the method proposed by  Colombari et al.~\cite{colombari2006bic}, each frame is divided
into blocks of size $N \times N$ overlapping by 50\% in both dimensions.
These blocks are clustered using single linkage agglomerative clustering along their time-line.
In the following step the background is built iteratively by selecting the best continuation block
for the current background using the principles of visual grouping. The
spatial correlations that naturally exist within small regions of the background image are considered
during the estimation process.
The algorithm can have problems with blending of the foreground and background
due to slow moving or quasi-stationary objects.
Furthermore, the algorithm is unlikely to achieve real-time performance due to its complexity.

\subsection{Hybrid Approaches}
\label{subsec_hybrid approaches}

In the third category, the algorithm presented by Gutchess et al.~\cite{gutchess2001bmi} has two stages.
The first stage is similar to that of~\cite{long1990sbg},
with the second stage estimating the likelihood of background visibility
by computing the optical flow of blocks between successive frames.
The motion information helps classify an intensity transition as background to foreground or vice versa.
The results are typically good, but the usage of optical flow for each pixel makes it computationally intensive.

In~\cite{cohen2005bel}, Cohen views the problem of estimating the background as an optimal labelling problem.
The method defines an energy function which is minimised to achieve an optimal solution at each pixel location.
It consists of \textit{data} and  \textit{smoothness} terms.
The data term accounts for pixel stationarity and motion boundary consistency
while the smoothness term looks for spatial consistency in the neighbourhood.
The function is minimised using the \textit{$\alpha$--expansion} algorithm~\cite{10.1109/ICCV.1999.791245} with suitable modifications.
A similar approach with a different energy function is proposed by Xu and Huang~\cite{xu2008lbp}.
The function is minimised using loopy belief propagation algorithm.
Both solutions provide robust estimates, however, their main drawback is large computational complexity
to process a small number of input frames.
For instance, in~\cite{xu2008lbp} the authors report a prototype of the algorithm on Matlab takes about 2.5 minutes
to estimate the background from a set of only 10 images of QVGA resolution {($320\times240$)}.

\section{Proposed Algorithm}
\label{sec:Proposed Algorithm}

\begin{figure*}[!t]
\begin{center}
  \begin{minipage}{1.0\textwidth}
    \includegraphics[width=0.23\columnwidth]{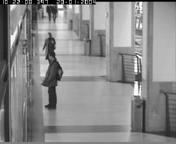}
    \hfill
    \includegraphics[width=0.23\columnwidth]{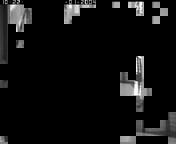}
    \hfill
    \includegraphics[width=0.23\columnwidth]{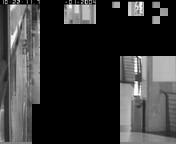}
    \hfill
    \includegraphics[width=0.23\columnwidth]{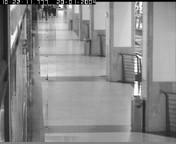}
  \end{minipage}

  ~

  \begin{minipage}{1.0\textwidth}
    \begin{minipage}{0.23\columnwidth}
      \centerline{\bf(i)}
    \end{minipage}
    \hfill
    \begin{minipage}{0.23\columnwidth}
      \centerline{\bf(ii)}
    \end{minipage}
    \hfill
    \begin{minipage}{0.23\columnwidth}
      \centerline{\bf(iii)}
    \end{minipage}
    \hfill
    \begin{minipage}{0.23\columnwidth}
      \centerline{\bf(iv)}
    \end{minipage}
  \end{minipage}

  \caption
    {
    \small
    {\bf (i)}~Example frame from an image sequence,
    {\bf (ii)}~partial background initialisation {(after~\textsl{Stage~2})},
    {\bf (iii)}~remaining background estimation in progress (\textsl{Stage 3}),
    {\bf (iv)}~estimated background.
    }
  \label{fig:BGI_iterations}
\end{center}
\end{figure*}

We propose a computationally efficient, region-level algorithm that aims to address the problems described in the previous section.
It has several additional advantages as well as novelties, including:

\begin{itemize}

\item
The background estimation problem is recast into an MRF scheme,
providing a theoretical framework.

\item
Unlike the techniques mentioned in Section~\ref{sec:Previous work},
it does not expect all frames of the sequence to be stored in memory simultaneously ---
instead, it processes frames sequentially, which results in a low memory footprint.

\item
The formulation of the clique potential in the MRF scheme
is based on the combined frequency response of the candidate block and its neighbourhood.
It is assumed that the most appropriate configuration results in the smoothest response (minimum energy),
indirectly exploiting the spatial correlations within small regions of a scene.

\item
Robustness against high frequency image noise.
In the calculation of the energy potential we compute 2D Discrete Cosine Transform (DCT) of the clique.
The high frequency DCT coefficients are ignored in the analysis as they typically represent image noise.

\end{itemize}

\subsection{Overview of the Algorithm}
\label{subsec:Proposed Algorithm_Overview}

In the text below we first provide an overview of the proposed algorithm,
followed by a detailed description of its components
(Sections~\ref{subsec:Similarity criteria for labels}~to~\ref{subsec:Calculation of energy potential}).
It is assumed that at each block location:
{\bf (i)}~the background is static and is revealed at some point in the training sequence for a short interval,
and
{\bf (ii)}~the camera is stationary.
The background is estimated by recasting it as a labelling problem in an MRF framework.
The algorithm has three stages.

Let the resolution of the greyscale image sequence {$I$} be {$\mathcal{W} \times \mathcal{H}$}.
In the first stage, the frames are viewed as instances of an undirected graph,
where the nodes of the graph are blocks of size {$N \times N$} pixels%
\footnote
  {
  For implementation purposes, each block location and its instances at every frame are treated as a node and its labels, respectively.
  }%
. We denote the nodes of the graph by {$\mathcal{N}{(i,j)}$}
for \mbox{$i = 0,1,2,\cdots,(\mathcal{W}/N) - 1$},  $j = 0,1,2,\cdots,(\mathcal{H}/N) - 1$.
Let {$I_f$} be the \mbox{$f$-th} frame of the training image sequence
and let its corresponding node labels be denoted by {$\mathcal{L}_f(i,j)$},
and {$f = 1,2,\cdots,F$},
where $F$ is the total number of frames.
For convenience, each node label {$\mathcal{L}_f(i,j)$} is vectorised
into an {$N^2$} dimensional vector {$\textbf{l}_f(i,j)$}.

At each node location {$(i,j)$},
a representative set {$\mathcal{R}(i,j)$} is maintained.
It contains distinct labels that were obtained along its temporal line.
Two labels are considered as distinct (visually different), if they
fail to adhere to one of the constraints described in Section~\ref{subsec:Similarity criteria for labels}.
Let these unique representative labels be denoted by 
{${\textbf{r}}_k(i,j)$} for {$k = 1, 2, \cdots, S$} (with {$S \leq F$}),
where {${\textbf{r}}_k$} denotes the mean of all the labels which 
were considered as similar to each other (mean of the cluster). 
Each label {${\textbf{r}}_k$} has an associated weight {$W_{k}$} which denotes its number of occurrences in the sequence,
i.e.,~the number of labels at location $(i,j)$ which are deemed to be the same as {${\textbf{r}}_k(i,j)$}.
For every such match, the corresponding {${\textbf{r}}_k(i,j)$}
and its associated variance, ${\mathbf \Sigma}_k(i,j)$
are updated recursively as given below:

\begin{equation}
{\textbf{r}}_{k}^{new} = {\textbf{r}}_{k}^{old} +  \frac{1}{W_k+1}\left(\textbf{l}_{f} - {\textbf{r}}_{k}^{old}\right)
\label{eqn:recursive_mean}
\end{equation}

\begin{equation}          
{\mathbf \Sigma}_k^{new} = \frac{W_k-1}{W_k} {\mathbf \Sigma}_k^{old} + 
\frac{1}{W_k+1}(\textbf{l}_{f} - {\textbf{r}}_{k}^{old})' (\textbf{l}_{f} - {\textbf{r}}_{k}^{old})
\label{eqn:recursive_var}
\end{equation}

\noindent where {${\textbf{r}}_{k}^{old}$, ${\mathbf \Sigma}_k^{old}$} and {${\textbf{r}}_{k}^{new}$}, ${\mathbf \Sigma}_k^{new}$
are the values of ${\textbf{r}}_{k}$ and its associated variance before and after
the update respectively, and $\textbf{l}_{f}$ is the incoming label
which matched ${\textbf{r}}_{k}^{old}$.
It is assumed that one element of {$\mathcal{R}(i,j)$} corresponds to the background.

In the second stage, representative sets {$\mathcal{R}(i,j)$}  having just one label
are used to initialise the corresponding node locations {$\mathcal{B}(i,j)$}
in the background {$\mathcal{B}$}.

In the third stage, the remainder of the background is estimated iteratively.
An optimal labelling solution is calculated by considering the likelihood of each of its labels
along with the \textit{a~priori} knowledge of the local spatial neighbourhood modelled as a MRF.
Iterated conditional mode (ICM), a deterministic relaxation technique, performs the optimisation.
The framework is described in detail in Section~\ref{subsec:MAP-MRF Framework}.
The strategy for selecting the location of an empty background node to initialise a label
is described in Section~\ref{subsec:Neighbourhood selection}.
The procedure for calculating the energy potentials, a prerequisite in determining the \textit{a~priori} probability,
is described in Section~\ref{subsec:Calculation of energy potential}.

The overall pseudo-code of the algorithm is given in Algorithm~1
and an example of the algorithm in action is shown in Figure~\ref{fig:BGI_iterations}.

\begin{figure*}[!t]
  \renewcommand{\baselinestretch}{1.2}
  \hrule
  \centerline{~}
  \begin{small}
    \emph{Stage 1: Collection of Label Representatives}
    \begin{enumerate}
      \item $ \mathcal{R}  \leftarrow \emptyset $ (null set)
      \item \textbf{for} $f = 1$ to $F$ \textbf{do}

      \begin{enumerate}
        \item Split input frame $I_f$ into node labels, each with a size of $N \times N$.
        \item {\bf for each} node label $\mathcal{L}_{f}(i,j)$ {\bf do}

        \begin{enumerate}

          \item
          Vectorise node $\mathcal{L}_{f}(i,j)$ into  $\textbf{l}_{f}(i,j)$.

          \item
          Find the representative label ${\textbf{r}}_{m}(i,j)$ from the set\\
          ${\mathcal{R}}(i,j) = \left({{\textbf{r}}_{k}(i,j)  | 1 \leq k \leq S}\right)$,
          matching to $\textbf{l}_{f}(i,j)$ \\
          based on  conditions in  Eqns.~(\ref{corr_ptrk}) and (\ref{diff_ptrk}).

          \begin{algorithmic}
          \IF{$($${\mathcal{R}}(i,j)$ = \{$\emptyset$\} or there is no match$)$}  
          \STATE {$k \leftarrow k + 1$.\\
          Add a new representative label ${\textbf{r}}_{k}(i,j)  \leftarrow ~\textbf{l}_{f}(i,j)$
          to set ${\mathcal{R}}(i,j)$ and initialise its weight, $W_{k}(i,j)$, to 1.}
          \ELSE
          \STATE{Recursively update the matched label ${\textbf{r}}_{m}(i,j)$ and its variance \\
    given by Eqns.~(\ref{eqn:recursive_mean})~and~(\ref{eqn:recursive_var}) respectively.\\
        $W_{m}(i,j) \leftarrow W_{m}(i,j) + 1 $ }
          \ENDIF
          \end{algorithmic}

        \end{enumerate}
        {\bf end for each}
      \end{enumerate}
    \textbf{end for}
    \end{enumerate}

    ~ \vspace{1.2ex}

    \emph{Stage 2: Partial Background Initialisation}
    \begin{enumerate}

      \item
      $\mathcal{B} \leftarrow \emptyset  $

      \item
      {\bf for each} set ${\mathcal{R}}\left(i,j\right)$  {\bf do}
      \begin{algorithmic}
      \IF{(size$\left({\mathcal{R}}\left(i,j\right) ) = 1\right) $}
      \STATE $\mathcal{B}\left(i,j\right) \leftarrow \textbf{r}_1\left(i,j\right).$
      \ENDIF
      \end{algorithmic}
      {\bf end for each}
     \end{enumerate}

    ~ \vspace{1.2ex}

    \emph{Stage 3: Estimation of the Remaining Background}
  \begin{enumerate}
      \item
     \textsl{ \small {Full background initialisation}}
    \begin{algorithmic}
      \WHILE{($\mathcal{B}$ not filled)}
      \IF
        {
        $\mathcal{B}\left(i,j\right) = \emptyset$ and has neighbours as specified in Section~\ref{subsec:Neighbourhood selection}
        }
      \STATE
        $\mathcal{B}\left(i,j\right) \leftarrow$ ${\textbf{r}}_{max}(i,j)$,
          the label out of set ${\mathcal{R}}\left(i,j\right)$
          which yields maximum value of the posterior probability described in Eqn.~(\ref{eq:bayestheoremlog})
          (see Section~\ref{subsec:MAP-MRF Framework}).
      \ENDIF
      \ENDWHILE
    \end{algorithmic}

  \item
       \textsl{ \small {Application of ICM}}

  \hspace{1pt}  $iteration\_count  \leftarrow 0$
        \begin{algorithmic}
 \WHILE{$(iteration\_count < total\_iterations$)}
    \FOR{\textbf{each} set ${\mathcal{R}}\left(i,j\right)$}
          \IF{$P(\textbf{r}_{new}{\footnotesize\left(i,j\right)}) > P(\textbf{r}_{old}{\footnotesize\left(i,j\right)})$}
          \STATE $\mathcal{B}\left(i,j\right) \leftarrow \textbf{r}_{new}\left(i,j\right)$,
            where $P(\cdot)$ is the posterior probability defined by Eqn.~(\ref{eq:bayestheoremlog}).
          \ENDIF
 \ENDFOR
\hspace{2pt}{\textbf{each}}

$iteration\_count = iteration\_count + 1$
          \ENDWHILE
        \end{algorithmic}

      \end{enumerate}

  \end{small}
  \vspace{1ex}
  \hrule
  \vspace{1ex}
    {
    \small
    Algorithm 1. Pseudo-code for the proposed algorithm.  A C++ implementation is available at \href{http://arma.sourceforge.net/background\_est/}{\bf http://arma.sourceforge.net/background\_est/}
    }
  \label{fig:pseudo-code}
  \renewcommand{\baselinestretch}{1.0}

  ~

  ~

  ~

  ~

  ~
\end{figure*}

\subsection{Similarity Criteria for Labels}
\label{subsec:Similarity criteria for labels}

We assert that two labels ${\textbf{l}}_f(i,j)$ and ${\textbf{r}}_k(i,j)$ are similar
if the following two constraints are satisfied:
\begin{equation}
  \frac
    {
    \left( {\textbf{r}}_{k}(i,j) - {\mu_{r_k}}(i,j) \right)' \left( {\textbf{l}}_{f}(i,j) - {\mu_{l_f}}(i,j) \right)
    }
    {
    %{\sigma_{r_k} \sigma_{l_t}}
    {\sigma_{r_k} \sigma_{l_f}}
    }
    > {\mathcal{T}}_{1}
\label{corr_ptrk}
\end{equation}%

\noindent
and
\begin{equation}
  \frac{1}{N^2} \sum\nolimits_{n = 0}^{N^2 - 1} \left|{\textit{d}}_{k_n}(i,j)\right|    < {\mathcal{T}}_2
  \label{diff_ptrk}
\end{equation}

\noindent
Equations~(\ref{corr_ptrk}) and (\ref{diff_ptrk}), respectively,
evaluate the correlation coefficient and the mean of absolute differences (MAD) between the two labels,
with the latter constraint ensuring that the labels are close in {$N^2$} dimensional space.
$\mu_{r_k},\mu_{l_f}$ and $\sigma_{r_k},\sigma_{l_f}$
are the mean and standard deviation of the elements of labels $\textbf{r}_{k}$ and $\textbf{l}_f$ respectively,
while {${\textbf{d}}_k(i,j) = {\textbf{l}}_f(i,j) - {\textbf{r}}_k(i,j)$}.

{$\mathcal{T}_1$}~is selected empirically (see Section~\ref{sec:Experimental Results}),
to ensure that two visually identical labels are not treated as being different due to image noise.
{$\mathcal{T}_2$}~is proportional to image noise and is found automatically as follows.
Using a short training video,
the MAD between co-located labels of successive frames is calculated.
Let the number of frames be {$L$} and {$N_b$} be the number of labels per frame.
The total MAD points obtained will be {$(L-1)N_b$}.
These points are sorted in ascending order and divided into quartiles.
The points lying between quartiles {$Q_3$} and {$Q_1$} are considered.
Their mean, {$\mu_{{\textsc{q}}_{31}}$}
and standard deviation, {$\sigma_{{\textsc{q}}_{31}}$},
are used to estimate {$\mathcal{T}_2$} as {$2 \times(\mu_{{\textsc{q}}_{31}} + 2\sigma_{{\textsc{q}}_{31}})$}.
This ensures that low MAD values (close or equal to zero)
and high MAD values (arising due to movement of objects) are ignored
(i.e.~treated as outliers).

We note that both constraints (\ref{corr_ptrk}) and (\ref{diff_ptrk}) are necessary.
As an example, two vectors {$[1,2,\cdots,16]$} and {$[101,102,\cdots,116]$}
have a perfect correlation of 1 but their MAD will be higher than {$\mathcal{T}_2$}.
On the other hand, if a thin edge of the foreground object is contained in one of the labels,
their MAD may be well within {$\mathcal{T}_2$}.
However, Eqn.~(\ref{corr_ptrk}) will be low enough to indicate the dissimilarity of the labels. In contrast,
we note that in~\cite{colombari2006bic} the similarity criteria is just based on the sum of squared distances
between the two blocks.

\begin{figure}[!t]
  \centerline{\includegraphics[width=0.5\columnwidth]{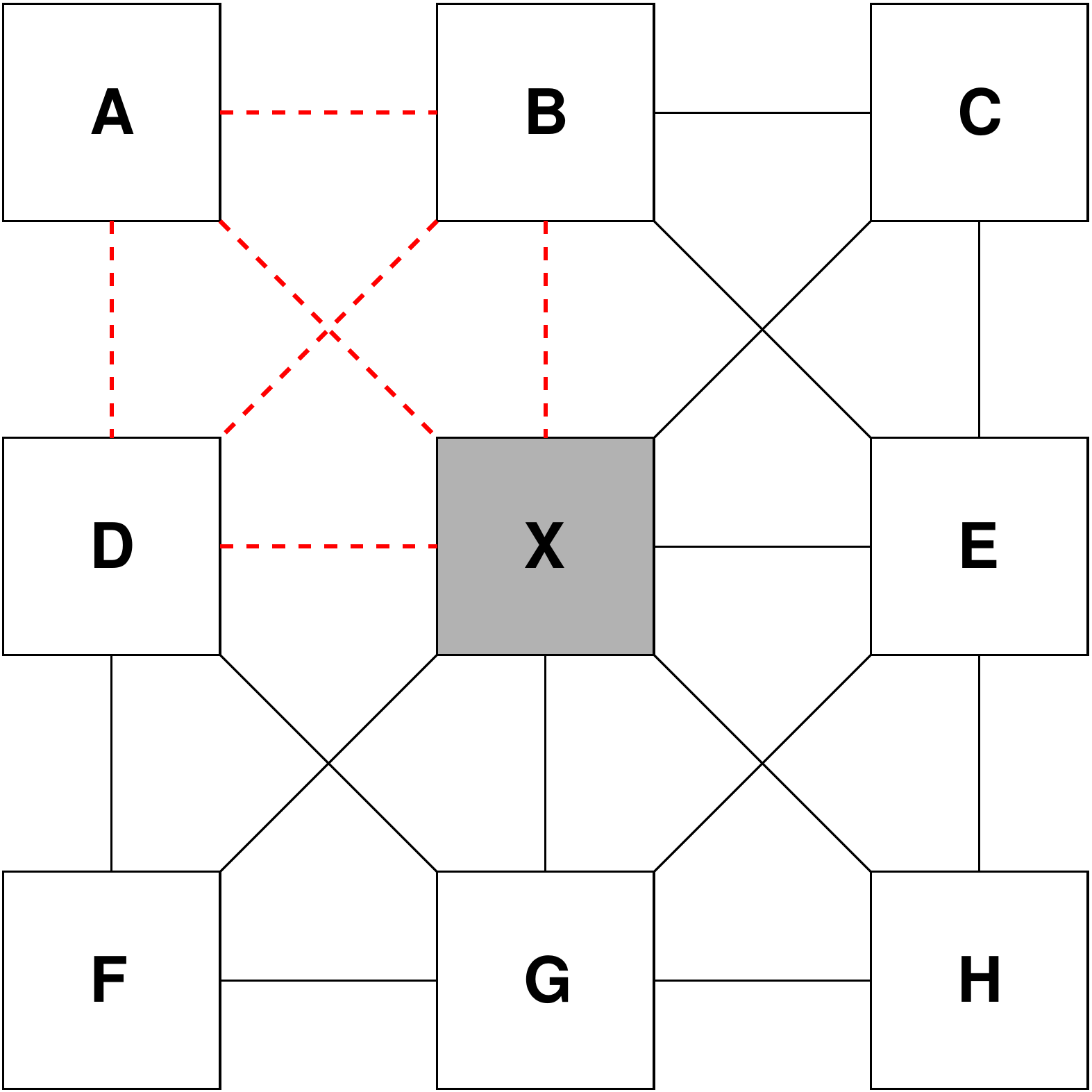}}
  \caption
    {
    \small
    The local neighbourhood system and its four cliques.
    Each clique is comprised of 4 nodes (blocks).
    To demonstrate one of the cliques, the the top-left clique has dashed links.
    }
  \label{fig:clique1}
\end{figure}

\begin{figure}[!t]
\begin{center}
\begin{minipage}{\columnwidth}
  {
  \begin{minipage}{.45\columnwidth}
    {
    \begin{minipage}{1\columnwidth}
      \centerline{\includegraphics[width=.4\columnwidth]{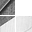}}
    \end{minipage}

    \vspace{15pt}
    \begin{minipage}{1\columnwidth}
      \centerline{\includegraphics[width=.4\columnwidth]{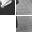}}
    \end{minipage}

    \vspace{15pt}
    \begin{minipage}{1\columnwidth}
      \centerline{\includegraphics[width=.4\columnwidth]{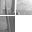}}
    \end{minipage}

    \vspace{15pt}
    \begin{minipage}{1\columnwidth}
      \centerline{\bf(i)}
    \end{minipage}
    }
  \end{minipage}
  \hfill
  \begin{minipage}{.45\columnwidth}
    {
    \begin{minipage}{1\columnwidth}
      \centerline{\includegraphics[width=.4\columnwidth]{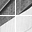}}
    \end{minipage}

    \vspace{15pt}
    \begin{minipage}{1\columnwidth}
      \centerline{\includegraphics[width=.4\columnwidth]{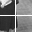}}
    \end{minipage}

    \vspace{15pt}
    \begin{minipage}{1\columnwidth}
      \centerline{\includegraphics[width=.4\columnwidth]{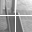}}
    \end{minipage}

    \vspace{15pt}
    \begin{minipage}{1\columnwidth}
      \centerline{\bf(ii)}
    \end{minipage}
    }
  \end{minipage}
  }
\end{minipage}

\caption
  {
  \small
  {\bf(i)}~Three cliques each of which has an empty node.
  The gaps between the blocks are for ease of interpretation only.
  {\bf(ii)}~Same cliques where the empty node has been labelled.
  The constraint of 3 neighbouring nodes to be available in 3 different directions
  as illustrated ensures that arbitrary edge continuities are taken into account
  while assigning the label at the empty node.
  }
\label{fig:edges}

\end{center}
\end{figure}

\subsection{Markov Random Field (MRF) Framework}
\label{subsec:MAP-MRF Framework}

Markov random field/probabilistic undirected graphical model theory
provides a coherent way of modelling context-dependent entities such as pixels or edges of an image.
It has a set of nodes, each of which corresponds to a variable or a group of variables,
and set of links each of which connects a pair of nodes.
In the field of image processing it has been widely employed to address many problems
that can be modelled as labelling problem with contextual information~\cite{geman1984srg,besag1986sad}. 

Let {$\textbf{X}$} be a 2D random field,
where each random variate {$\textit{X}_{(i,j)}$ $( \forall \hspace{2pt}i,j )$}
takes values in discrete \textit{state space} {$\Lambda$}.
Let {$\omega$} {$\in$} {$\Omega$} be a \textit{configuration}
of the variates in {$\textbf{X}$}, and let {$\Omega$} be the set of all such configurations.
The joint probability distribution of  $\textbf{X}$ is considered Markov if
\begin{equation}
 {  p(\textbf{X} = \omega) > 0, \hspace{2pt} \forall \hspace{3pt}\omega \in \Omega }
   \label{eq:mrf_conds1}
\end{equation}%

\noindent
and
\begin{equation}
{
    p\left(\textit{X}_{(i,j)} | \textit{X}_{(p,q)}, {\textsl{\footnotesize{(i, j)}} \neq \textsl{\footnotesize{(p, q)}}}\right) \\
    = p\left(\textit{X}_{(i,j)} | \textit{X}_{{N}_{(i,j)}}\right)}
\label{eq:mrf_conds1b}
\end{equation}%

\noindent
where {$\textit{X}_{{N}_{(i,j)}}$} refers to the local \textit{neighbourhood system} of {$\textit{X}_{(i,j)}$}.

Unfortunately, the theoretical factorisation of the joint probability distribution of the MRF
turns out to be intractable.
To simplify and provide computationally efficient factorisation,
Hammersley-Clifford theorem~\cite{besag1974sia}
states that an MRF can equivalently be characterised by a Gibbs distribution.
Thus

\begin{equation}
{
    p(\textbf{X}= \omega) = \frac{e^{-U(\omega)/T}}{Z}}
    \label{eq:mrf_defines1}
\end{equation}%

\noindent
where
\begin{equation}
{
    Z = \sum\nolimits_{\omega}{e^{-U(\omega)/T}}}
    \label{eq:mrf_defines2}
\end{equation}
is a normalisation constant known as the \textit{partition function},
{$T$} is a constant used  to moderate the peaks of the distribution
and {$U(\omega)$} is an \textit{energy function}
which is the sum of \textit{clique/energy potentials} {$V_c$} over all possible cliques~$C$:
\begin{equation}
{   U(\omega) = \sum\nolimits_{c \in \textit{C}}V_c(\omega)    }
    \label{eq:mrf_defines3}
\end{equation}

\noindent
The value of {$V_c(\omega)$} depends on the local configuration of clique~$c$.

In our framework, information from two disparate sources is combined using Bayes' rule.
The local visual observations at each node to be labelled yield label likelihoods.
The resulting label likelihoods are combined with \textit{a~priori} spatial knowledge of the neighbourhood represented as an MRF.

Let each input image {$I_f$} be treated as a realisation of the random field {$\mathcal{B}$}.
For each node {$\mathcal{B}(i,j)$}, the  representative set {$\mathcal{R}(i,j)$}
(see Section~\ref{subsec:Proposed Algorithm_Overview})
containing unique labels is treated as its \textit{state space} with each {$\textbf{r}_k(i,j)$} as its plausible label%
\footnote{To simplify the notations, index term $(i,j)$ has been henceforth omitted.}%
.

Using  Bayes' rule, the posterior probability for every label at each node is derived
from the \textit{a~priori} probabilities and the observation-dependent likelihoods given by
\begin{equation}
   {P(\textbf{r}_{k}) = l(\textbf{r}_{k})p(\textbf{r}_{k})}
\label{eq:bayestheorem1}
\end{equation}

The product is comprised of  likelihood {$l(\textbf{r}_k)$} of each label
{$\textbf{r}_{k}$} of set {$\mathcal{R}$} and its \textit{a~priori} probability density
{$ p(\textbf{r}_{k})$}, conditioned on its local neighbourhood.
In the derivation of likelihood function it is assumed that at each node the observation components {$\textbf{r}_k$}
are conditionally independent and have the same known conditional density function dependent only on that node.

At a given node, the label that yields maximum \textit{a posteriori} (MAP) probability
is chosen as the best continuation of the background at that node.

To optimise the MRF-based function defined in Eqn.~(\ref{eq:bayestheorem1}),
ICM is used since it is computationally efficient and avoids large scale effects%
\footnote{An undesired characteristic where a single label wrongly gets assigned to most of the nodes of the random field.}%
~\cite{besag1986sad}.
ICM  maximises local conditional probabilities iteratively until convergence is achieved.

Typically, in ICM an initial estimate of the labels is obtained by maximising the likelihood function.
However, in our framework an initial estimate consists of partial reconstruction of the background
at nodes having just one label which is assumed to be the background.
Using the available background information, the remaining unknown background is estimated progressively
(see Section~\ref{subsec:Neighbourhood selection}).

At every node, the likelihood of each of its labels {$\textbf{r}_k$ $(k = 1, 2, \cdots, S)$}
is calculated using corresponding weights {$W_k$} (see Section~\ref{subsec:Proposed Algorithm_Overview}).
The higher the occurrences of a label, the more is its likelihood to be part of the background.
Empirically, the likelihood function is modelled by a simple weighted function given by:
\begin{equation}
 l(\textbf{r}_k)= \frac{W_{c_k}}{\sum_{k=1}^{S}{W_{c_k}}}
    \label{eq:rayleigh}
\end{equation}%

\noindent
where {$W_{c_k}$ = min$(W_{max},W_k)$} and $W_{max}$ = 5 $\times$ frame rate of the captured sequence%
\footnote
  {
  It is assumed that the likelihood of a label exposed for a duration of 5 seconds
  is good enough  to be regarded as a potential candidate for the background.
  }%
.

As evident, the weight $W$ of a label greater than $W_{max}$ will be capped to $W_{max}$.
Setting a maximum threshold value is necessary in circumstances where the image sequence
has a stationary foreground object visible for an exceedingly long period
 when compared to the background occluded by it.
For example, in a 1000-frame sequence, a car might be parked for the first 950 frames and in the last 50 frames it drives away.
In this scenario, without the cap the likelihood of the car being part of the background
will be too high compared to the true background
and this will bias the overall estimation process causing errors in the estimated background.

Relying on this likelihood function alone is insufficient since it may still introduce estimation errors
even when the foreground object is exposed for just slightly longer duration compared to the background.

Hence, to overcome this limitation, the spatial neighbourhood modelled as Gibbs distribution
(given by Eqn.~(\ref{eq:mrf_defines1})) is encoded into an \textit{a~priori} probability density.
The formulation of the clique potential {$V_c(\omega)$} referred in Eqn.~(\ref{eq:mrf_defines3})
is described in the Section~\ref{subsec:Calculation of energy potential}.
Using  Eqns.~(\ref{eq:mrf_defines1}),~(\ref{eq:mrf_defines2})~and~(\ref{eq:mrf_defines3})
the calculated clique potentials $V_c(\omega)$ are transformed into \textit{a~priori} probabilities.
For a given label, the smaller the value of energy function, the
greater is its probability in being the best match with respect to its neighbours.

In our evaluation of the posterior probability given by Eqn.~(\ref{eq:bayestheorem1}),
the local spatial context term is assigned more weight than the likelihood function which is just based on temporal statistics.
Thus, taking log of Eqn.~(\ref{eq:bayestheorem1}) and assigning a weight to the prior, we get:

\begin{equation}
  {\log\left(P(\textbf{r}_{k})\right) =
  \log \left(l(\textbf{r}_{k})\right) + \eta \hspace{1pt}
  \log \left(p(\textbf{r}_{k})\right)}
\label{eq:bayestheoremlog}
\end{equation}%

\noindent
where $\eta$ has been empirically set to number of neighbouring nodes used in clique potential calculation
(typically $\eta$ = 3).

The weight is required in order to  address the scenario where the true background label
is visible for a short interval of time when compared to labels containing the foreground.
For example, in Figure~\ref{fig:BGI_iterations},
a sequence consisting of 450 frames was used to estimate its background.
The person was standing as shown in Figure~\ref{fig:BGI_iterations}(i)
for the first 350 frames and eventually walked off during the last 100 frames.
The algorithm was able to estimate the background occluded by the standing person.
It must be noted that pixel-level processing techniques are likely to fail in this case.

\subsection{Node Initialisation}
\label {subsec:Neighbourhood selection}

Nodes containing a single label in their representative set are directly initialised
with that label in the background (see Figure~\ref{fig:BGI_iterations}(ii)).
However, in some rare situations there is a possibility that all the sets may contain
more than one label. In such a case, the algorithm heuristically picks the label having the largest weight $W$
from the representative sets of the four corner nodes as an initial seed to initialise the background.
It is assumed atleast one of the corner regions in the video frames corresponds to a static region.

The rest of the nodes are initialised based on constraints as explained below.
In our framework, the local \textit{neighbourhood system}~\cite{geman1984srg} of a node and the corresponding cliques
are defined as shown in Figure~\ref{fig:clique1}.
A clique is defined as a subset of the nodes in the neighbourhood system that are fully connected.
The background at an empty node will be assigned only if at least 2 neighbouring nodes of its \mbox{4-connected} neighbours
adjacent to each other and the diagonal node located between them are already assigned with background labels.
For instance, in Figure~\ref{fig:clique1}, we can assign a label to node $X$ if at least nodes $B$, $D$ (adjacent 4-connected neighbours)
and $A$ (diagonal node) have already been assigned with labels.
In other words, label assignment at node $X$ is \textit{conditionally independent} of all other nodes given these 3 neighbouring nodes.

Node $X$ has nodes $D$, $B$, $E$ and $G$ as its 4-connected neighbours.
Let us assume that all nodes except $X$ are labelled.
To label node $X$ the procedure is as follows.
In Figure~\ref{fig:clique1}, four cliques involving $X$ exist.
For each candidate label at node $X$, the energy potential for each of the four cliques
is evaluated independently given by Eqn.~(\ref{eq:energypotential})
and summed together to obtain its energy value. The label that yields the least value is likely to be
assigned as the background.

Mandating that the background should be available in at least 3 neighbouring nodes
located in three different directions with respect to node $X$ ensures that the best match is obtained
after evaluating the continuity of the pixels in all possible orientations.
For example, in Figure~\ref{fig:edges}, this constraint ensures that the edge orientations are
well taken into account in the estimation process.
It is evident from examples in Figure~\ref{fig:edges} that using either horizontal or vertical neighbours alone
can cause errors in background estimation (particularly at edges).

Sometimes not all the three neighbours are available.
In such cases, to assign a label at node $X$ we use one of its 4-connected neighbours
whose node has already been assigned with a label.
Under these contexts, the clique is defined as two adjacent nodes either in the horizontal or vertical direction.

Typically, after initialising all the empty nodes an accurate estimate of the background is obtained.
Nonetheless, in certain circumstances an incorrect label assignment at a node may cause an error
to occur and propagate to its neighbourhood.
Our previous algorithm~\cite{vreddysbe2009} is prone to this type of problem.
However, in the current framework the problem is successfully redressed by the application of ICM.
In subsequent iterations, in order to avoid redundant calculations,
the label process is carried out only at nodes where a change in the label of one of their 8-connected neighbours
occurred in the previous iteration.

\begin{figure}[!t]

  \begin{center}
    \begin{minipage}{0.9\columnwidth}
      \begin{center}

 \begin{minipage}{1.0\columnwidth}
          \begin{center}
            \begin{minipage}{0.05\columnwidth}
              \centerline{\bf(i)}
            \end{minipage}
      \hspace{4pt}
            \begin{minipage}{0.15\columnwidth}
              \begin{center}
                {\includegraphics[width=1\columnwidth]{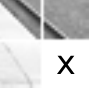}}
              \end{center}
            \end{minipage}
            ~
            \begin{minipage}{0.50\columnwidth}
              \begin{center}
                  {\footnotesize 1}~\includegraphics[width=0.15\columnwidth]{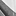} ~
                  {\footnotesize 2}~\includegraphics[width=0.15\columnwidth]{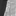}
                  \\~\\
                  {\footnotesize 3}~\includegraphics[width=0.15\columnwidth]{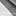} ~
                  {\footnotesize 4}~\includegraphics[width=0.15\columnwidth]{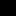} 
              \end{center}
            \end{minipage}
          \end{center}
        \end{minipage}

        ~

        \begin{minipage}{1.0\columnwidth}
          \begin{center}
              \hrule
          \end{center}
        \end{minipage}

%*************************************************************************

 \begin{minipage}{1.0\columnwidth}
          \begin{center}
            \begin{minipage}{0.05\columnwidth}
              \centerline{\bf(ii)}
            \end{minipage}
      \hspace{4pt}
            \begin{minipage}{0.15\columnwidth}
              \centerline{\includegraphics[width=\columnwidth]{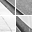}}
            \end{minipage}
            \begin{minipage}{0.70\columnwidth}
              \centerline{\includegraphics[width=\columnwidth]{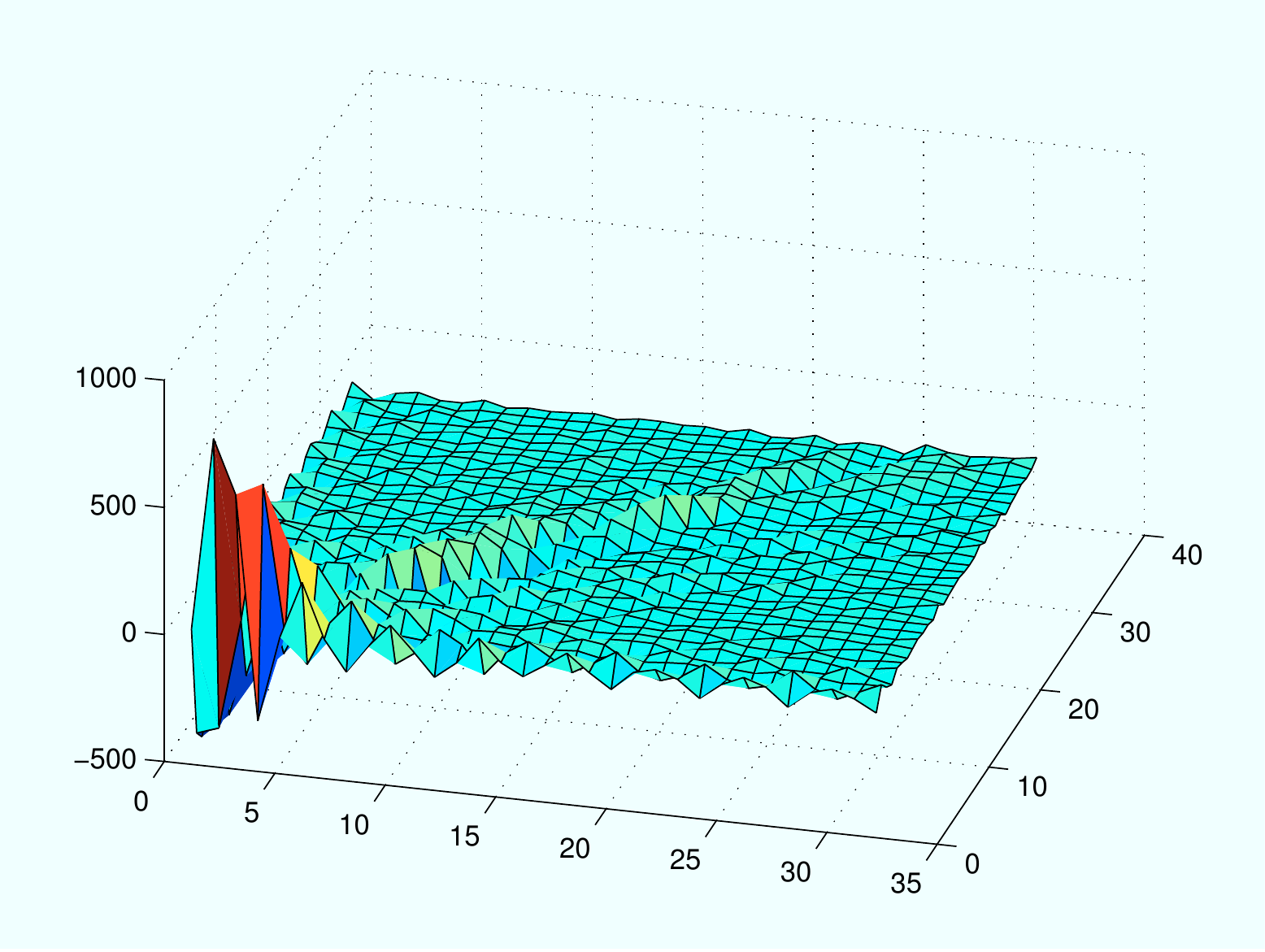}}
            \end{minipage}
          \end{center}
        \end{minipage}

        \begin{minipage}{1.0\columnwidth}
          \begin{center}
              \hrule
          \end{center}
        \end{minipage}

%*************************************************************************
 \begin{minipage}{1.0\columnwidth}
          \begin{center}
            \begin{minipage}{0.05\columnwidth}
              \centerline{\bf(iii)}
            \end{minipage}
      \hspace{4pt}
            \begin{minipage}{0.15\columnwidth}
              \centerline{\includegraphics[width=\columnwidth]{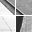}}
            \end{minipage}
            \begin{minipage}{0.70\columnwidth}
              \centerline{\includegraphics[width=\columnwidth]{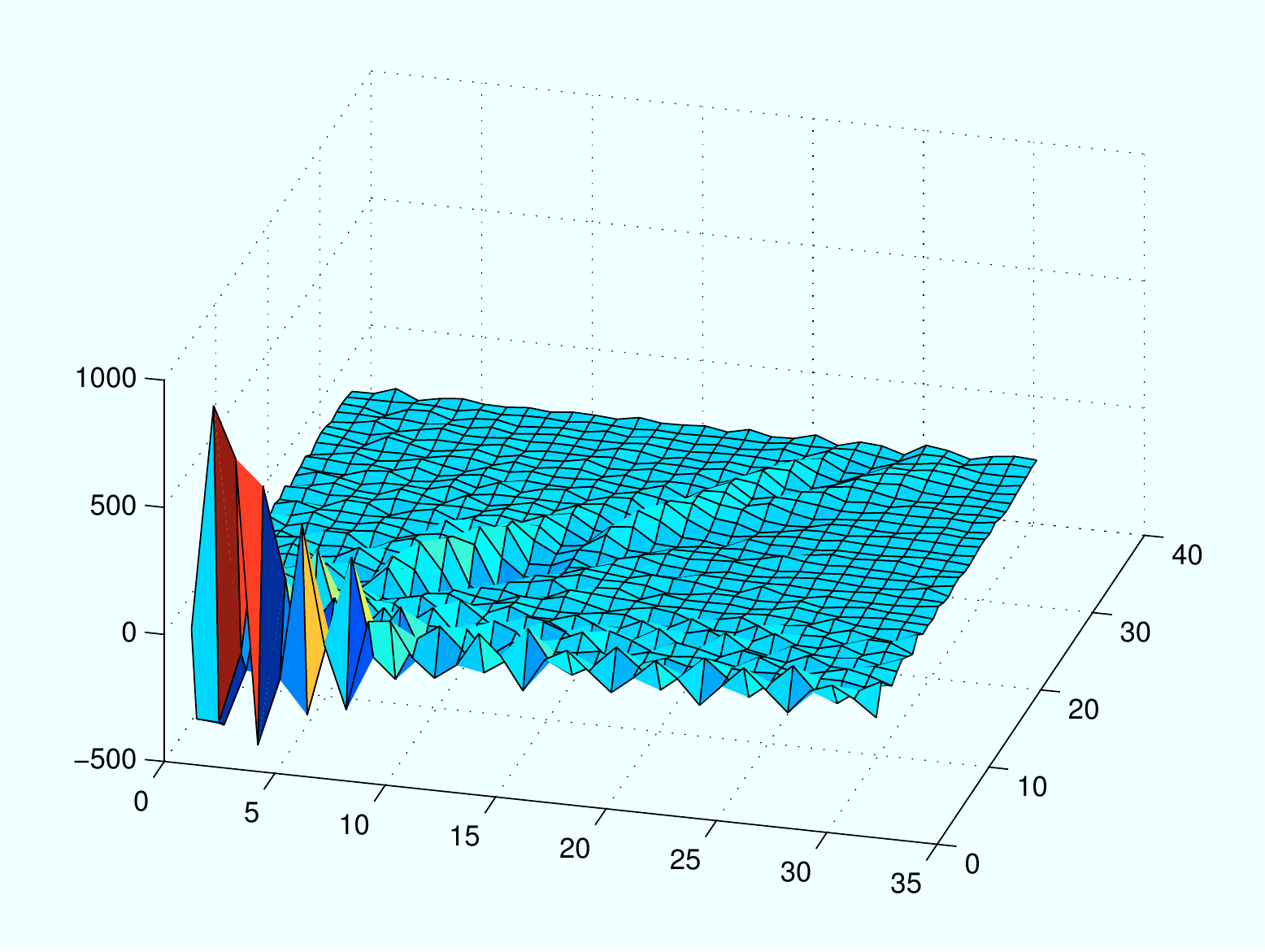}}
            \end{minipage}
          \end{center}
        \end{minipage}

        \begin{minipage}{1.0\columnwidth}
          \begin{center}
              \hrule
          \end{center}
        \end{minipage}
%*************************************************************************
        \begin{minipage}{1.0\columnwidth}
          \begin{center}
            \begin{minipage}{0.05\columnwidth}
              \centerline{\bf(iv)}
            \end{minipage}
      ~
            \begin{minipage}{0.15\columnwidth}
              \centerline{\includegraphics[width=1\columnwidth]{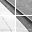}}
            \end{minipage}
            \begin{minipage}{0.70\columnwidth}
              \centerline{\includegraphics[width=\columnwidth]{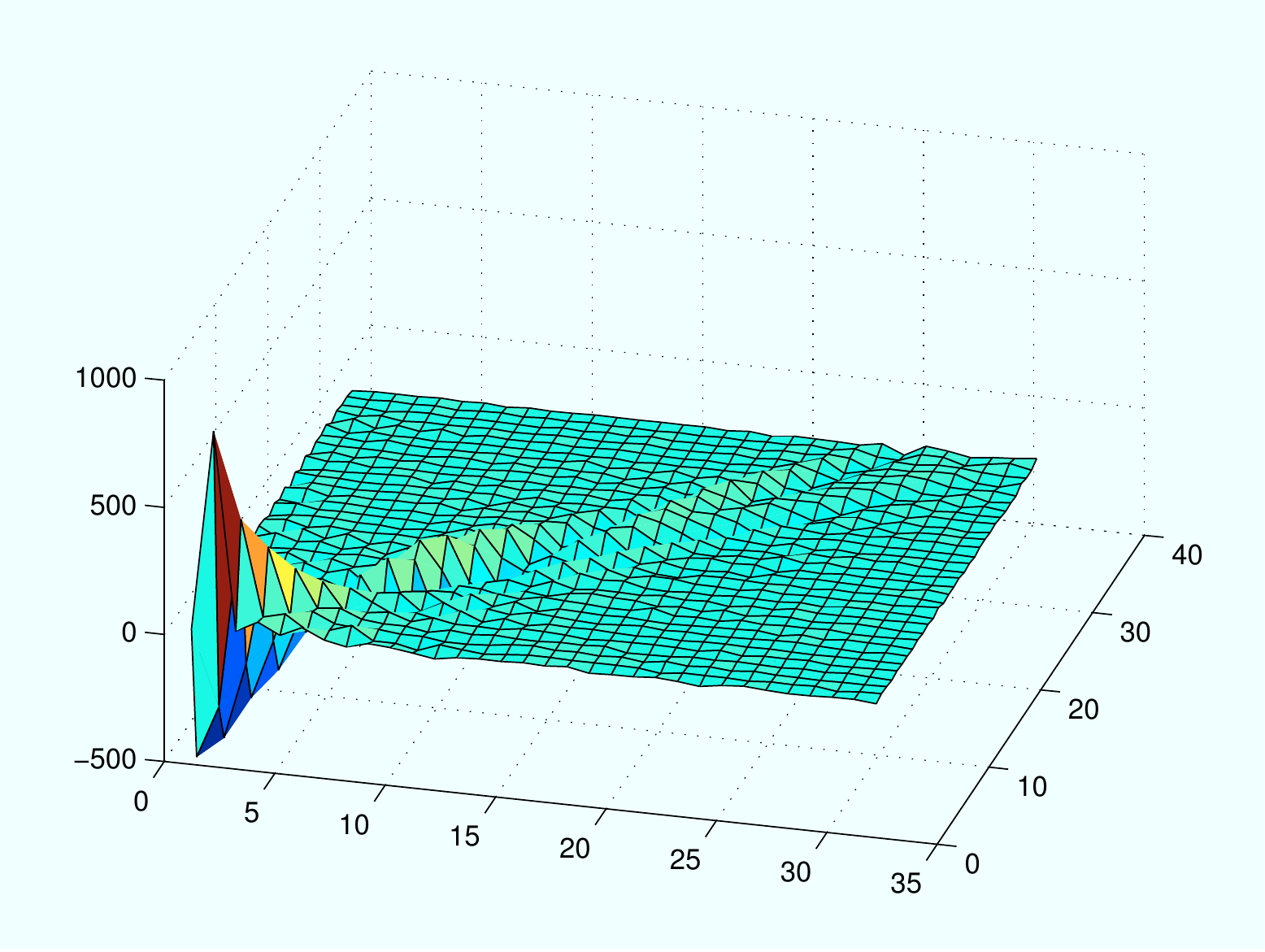}}
            \end{minipage}
          \end{center}
        \end{minipage}
        \begin{minipage}{1.0\columnwidth}
          \begin{center}
              \hrule
          \end{center}
        \end{minipage}

%*************************************************************************

   \begin{minipage}{1.0\columnwidth}
          \begin{center}
            \begin{minipage}{0.05\columnwidth}
              \centerline{\bf(v)}
            \end{minipage}
      \hspace{4pt}
            \begin{minipage}{0.15\columnwidth}
              \centerline{\includegraphics[width=1\columnwidth]{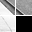}}
            \end{minipage}
            \begin{minipage}{0.70\columnwidth}
              \centerline{\includegraphics[width=0.90\columnwidth]{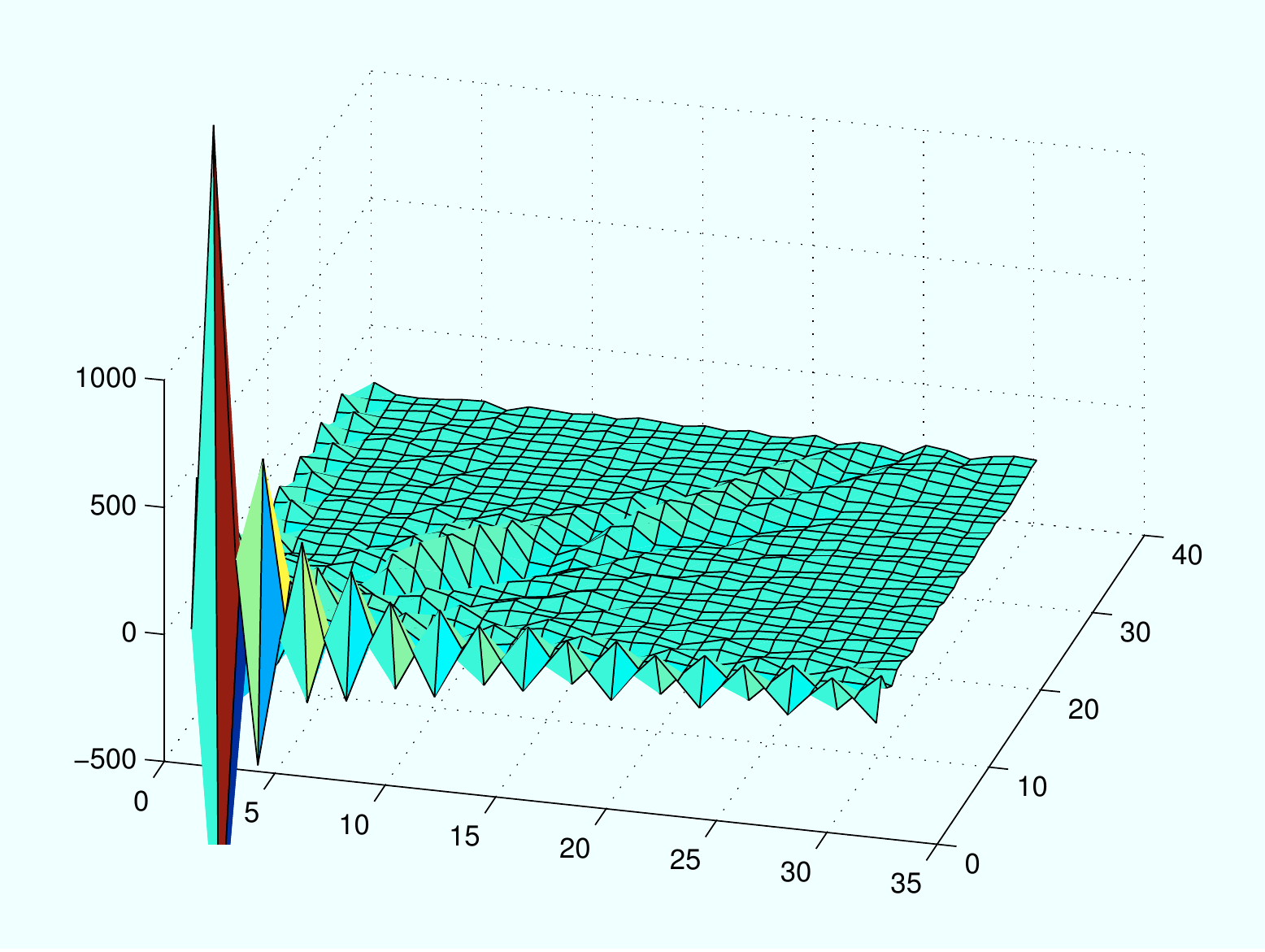}}
            \end{minipage}
          \end{center}
        \end{minipage}

        \begin{minipage}{1.0\columnwidth}
          \begin{center}
              \hrule
          \end{center}
        \end{minipage}

      \end{center}
    \end{minipage}

    \centerline{~}

    \caption
      {
      \small
      An example of the processing done in Section~\ref{subsec:Calculation of energy potential}.
      {\bf (i)}~A~clique involving empty node $X$ with four candidate labels in its representative set.
      {\bf (ii)}~A clique and a graphical representation of its DCT coefficient matrix
      where node $X$ is initialised with candidate label~1.
      The gaps between the blocks are for ease of interpretation only
      and are not present during DCT calculation. 
      {\bf(iii)}~As per~(ii), but using candidate label~2. 
      {\bf(iv)}~As per~(ii), but using candidate  label~3. 
      {\bf(v)}~As per~(ii), but using candidate label~4. 
      The smoother spectral distribution for candidate~3 suggests
      that it is a better fit than the other candidates.
      }
    \label{fig:FreqDist}

  \end{center}

  ~

  ~
\end{figure}

\subsection{Calculation of the Energy Potential}
\label{subsec:Calculation of energy potential}

In Figure~\ref{fig:clique1}, it is assumed that all nodes except {$X$} are assigned with the background labels.
The algorithm needs to assign an optimal label at node {$X$}. Let node {$X$} have {$S$} labels
in its state space {${\mathcal{R}}$} for {$k = 1, 2, \cdots, S$}
where one of them represents the true background.
Choosing the best label is accomplished by analysing the spectral response of every possible clique
constituting the unknown node $X$.
For the decomposition we chose the Discrete Cosine Transform (DCT)~\cite{ahmed1974dct}
due to its decorrelation properties as well as ease of implementation in hardware.
The DCT coefficients were also utilised by~Wang et al.~\cite{wang2008modeling}
to segment moving objects from compressed videos.

We consider the top left clique consisting of nodes $A$, $B$, $D$ and {$X$}.
Nodes $A$, $B$ and $C$ are assigned with background labels. 
Node {$X$} is assigned with one of {$S$} candidate labels.
We take the 2D DCT of the resulting clique.
The transform coefficients are stored in matrix $\textbf{C}_{k}$ of size \mbox{$M \times M$} (\mbox{$M = 2N$})
with its elements referred to as {$C_k{(v,u)}$}.
The term {$C_{k}{(0,0)}$} (reflecting the sum of pixels at each node) is forced to 0
since we are interested in analysing the spatial variations of pixel values.

Similarly, for other labels present in the state space of node {$X$}, we compute their corresponding 2D DCT as mentioned above.
A graphical example of the procedure is shown in Figure~\ref{fig:FreqDist}.

Assuming that pixels close together have similar intensities, 
When the correct label is placed at node {$X$}, 
the resulting transformation has a smooth response (less high frequency components) when compared to other 
candidate labels.

The higher-order components typically correspond to high frequency image noise.
Hence, in our energy potential calculation defined below we consider only the lower 75\% of the frequency components
after performing a zig-zag scan from the origin.

The energy potential for each label is calculated using:

\begin{equation}
{
  V_c(\omega_k) =
  \left(
    \sum\nolimits_{v = 0}^{P-1} \sum\nolimits_{u = 0}^{P-1} \left| C_k{(v,u)} \right|
  \right)}
\label{eq:energypotential}
\end{equation}%

\noindent
where {$P = \mbox{ceil}\left(\sqrt{M^2 \times 0.75}\right)$} and $\omega_k$ is the local configuration involving label $k$.
Similarly, the potentials over other three cliques in Figure~\ref{fig:clique1} are calculated.

\section{Experiments}
\label{sec:Experimental Results}

In our experiments the testing was limited to greyscale sequences.
The size of each node was set to { $16 \times 16$}.
The threshold $\mathcal{T}_1$ was empirically set to 0.8 based on preliminary experiments,
discussed in subsection~\ref{subsec:T1Sensitivity}.
{ $\mathcal{T}_2$} (found automatically) was found to vary between 1 and 4 when tested on several image sequences
({$\mathcal{T}_1$} and {$\mathcal{T}_2$} are described in Section~\ref{subsec:Similarity criteria for labels}).

A prototype of the algorithm using Matlab on a 1.6 GHz dual core processor yielded 17~fps.
We expect that considerably higher performance can be attained by converting the implementation to C++,
with the aid of libraries such as OpenCV~\cite{Bradski2008} or Armadillo~\cite{Armadillo_2010}.
To emphasise the effectiveness of our approach,
the estimated backgrounds were obtained by labelling all the nodes just once (no subsequent iterations were performed).

We conducted two separate set of experiments to verify the performance of the proposed method.
In the first case, we measured the quality of the estimated backgrounds,  
while in the second case we evaluated the influence of 
the proposed method on a foreground segmentation algorithm.
Details of both the experiments are described 
in Sections~\ref{sec_Standalone Evaluation} and~\ref{sec_Evaluation_by_Foreground_Segmentation},
respectively.
 
\subsection{Standalone Performance}
\label{sec_Standalone Evaluation}

We compared the proposed algorithm with a median filter based approach 
(i.e. applying filter on pixels at each location across all the frames)
as well as finding intervals of stable intensity (ISI) method presented in~\cite{wang2006nrs}.
We used a total of 20 surveillance videos:
7 obtained from CAVIAR dataset\footnote{http://groups.inf.ed.ac.uk/vision/CAVIAR/CAVIARDATA1/}, 
3 sequences from the abandoned object dataset used in the CANDELA project\footnote{http://www.multitel.be/{$\sim$}va/candela/}
and 10 unscripted sequences obtained from a railway station in Brisbane.
The CAVIAR and and CANDELA sequences were chosen based on four criteria:
{\bf (i)} a minimum duration of 700 frames,
{\bf (ii)} containing significant background occlusions,
{\bf (iii)} the true background is available in at least one frame,
and
{\bf (iv)} have largely static backgrounds. 
Having the true background allows for quantitative evaluation of the accuracy of background estimation.
The sequences were resized to $320 \times 240$ pixels (QVGA resolution)
in keeping with the resolution typically used in the literature.

The algorithms were subjected to both qualitative and quantitative evaluations.
Subsections~\ref{sec_Qualitative Evaluation} and~\ref{sec_Quantitative Evaluation} respectively 
describe the experiments for both cases.
Sensitivity of {$\mathcal{T}_1$} is studied in subsection~\ref{subsec:T1Sensitivity}.

\begin{figure*}[!t]

  \begin{minipage}{1\textwidth}
    \includegraphics[width=0.24\columnwidth]{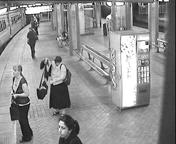}
    \hfill
    \includegraphics[width=0.24\columnwidth]{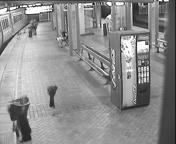}
    \hfill
    \includegraphics[width=0.24\columnwidth]{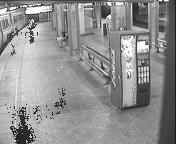}
    \hfill
    \includegraphics[width=0.24\columnwidth]{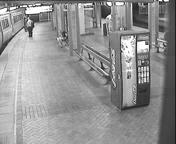}
  \end{minipage}
  
  \vspace{1ex}

  \begin{minipage}{1\textwidth}
    \includegraphics[width=0.24\columnwidth]{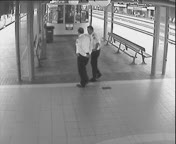}
    \hfill
    \includegraphics[width=0.24\columnwidth]{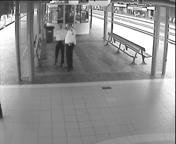}
    \hfill
    \includegraphics[width=0.24\columnwidth]{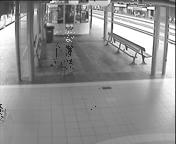}
    \hfill
    \includegraphics[width=0.24\columnwidth]{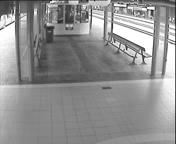}
  \end{minipage}
  
  \vspace{1ex}
  
  \begin{minipage}{1\textwidth}
    \includegraphics[width=0.24\columnwidth]{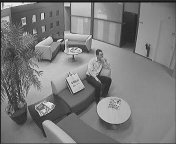}
    \hfill
    \includegraphics[width=0.24\columnwidth]{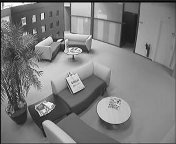}
    \hfill
    \includegraphics[width=0.24\columnwidth]{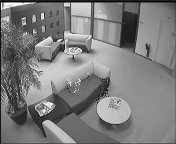}
    \hfill
    \includegraphics[width=0.24\columnwidth]{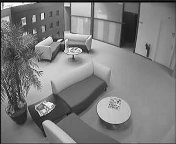}
  \end{minipage}
  
  \vspace{1ex}
  
  \begin{minipage}{1\textwidth}
    \includegraphics[width=0.24\columnwidth]{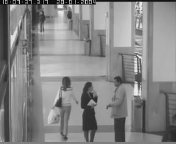}
    \hfill
    \includegraphics[width=0.24\columnwidth]{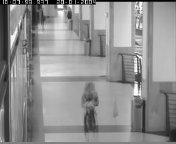}
    \hfill
    \includegraphics[width=0.24\columnwidth]{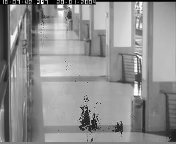}
    \hfill
    \includegraphics[width=0.24\columnwidth]{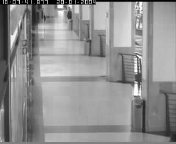}
  \end{minipage}

  \vspace{1ex}

  \begin{minipage}{1\textwidth}
    \begin{minipage}{0.24\columnwidth}
      \centerline{\bf(i)}
    \end{minipage}
    \hfill
    \begin{minipage}{0.24\columnwidth}
      \centerline{\bf(ii)}
    \end{minipage}
    \hfill
    \begin{minipage}{0.24\columnwidth}
      \centerline{\bf(iii)}
    \end{minipage}
    \hfill
    \begin{minipage}{0.24\columnwidth}
      \centerline{\bf(iv)}
    \end{minipage}
  \end{minipage}

  \vspace{1ex}

  \caption
    {
    \small
    {\bf (i)}~Example frames from four videos,
    and the reconstructed background using:
    {\bf (ii)}~median filter,
    {\bf (iii)}~ISI method~\cite{wang2006nrs},
    {\bf (iv)}~proposed method.
    }
  \label{results_comp123}  

\end{figure*}

\subsubsection{Qualitative Evaluation}
\label{sec_Qualitative Evaluation}

All 20 sequences were used for subjective evaluation of the quality of background estimation.
Figure~\ref{results_comp123} shows example results on four sequences with differing complexities.

Going row by row, the first and second sequences are from a railway station in Brisbane,
the  third is from the CANDELA dataset
and the last is from the CAVIAR dataset.
In the first sequence, several commuters wait for a train, slowly moving around the platform.
In the second sequence, two people (security guards) are standing on the platform for most of the time.
In the third sequence, a person places a bag on the couch, abandons it and walks away.
Later, the bag is picked up by another person.
The bag is in the scene for about 80\% of the time.
In the last sequence two people converse for most of the time while others slowly walk along the corridor.
All four sequences have foreground objects that are either dynamic or quasi-stationary for most of the time.

It can be observed that the estimated backgrounds obtained from  median filtering (second column)
and the ISI method (third column) have traces of foreground objects that were stationary for a relatively long time.
The results of the proposed method appear in the fourth column and indicate visual improvements over the other two techniques.
It must be noted that stationary objects can appear as background to the proposed algorithm,
as indicated in the first row of the fourth column.
Here a person is standing at the far end of the platform for the entire sequence.

\subsubsection{Quantitative Evaluation}
\label{sec_Quantitative Evaluation}

To objectively evaluate the quality of the estimated backgrounds
we considered the test criteria described in~\cite{gutchess2001bmi},
where the average grey-level error (AGE),
total number of error pixels (EPs)
and the number of ``clustered'' error pixels (CEPs) are used.
AGE is the average of the difference between the true and estimated backgrounds. 
If the difference between estimated and true background pixel is greater than a threshold,
then it is classified as an EP.
We set the threshold to 20, to ensure good quality backgrounds.
A~CEP is defined as any error pixel whose 4-connected neighbours are also error pixels.
As our method is based on region-level processing we calculated only the AGE and CEPs.

\begin{figure*}[!tb]
\begin{center}

  \begin{minipage}{0.4\textwidth}
    \centerline{\includegraphics[width=1\columnwidth]{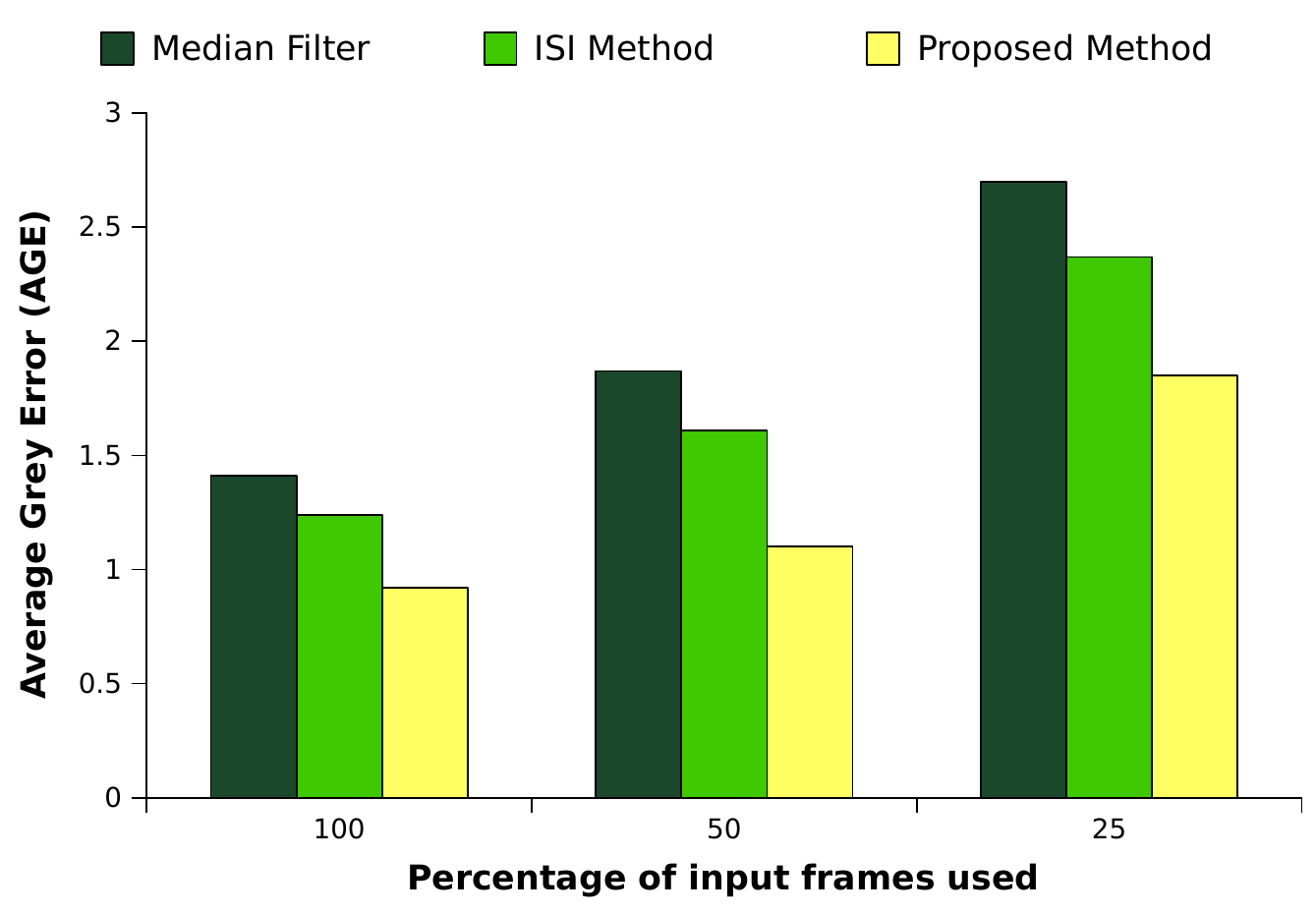}}
    \centerline{\bf (i)}
  \end{minipage}
  ~~~
  \begin{minipage}{0.4\textwidth}
    \centerline{\includegraphics[width=1\columnwidth]{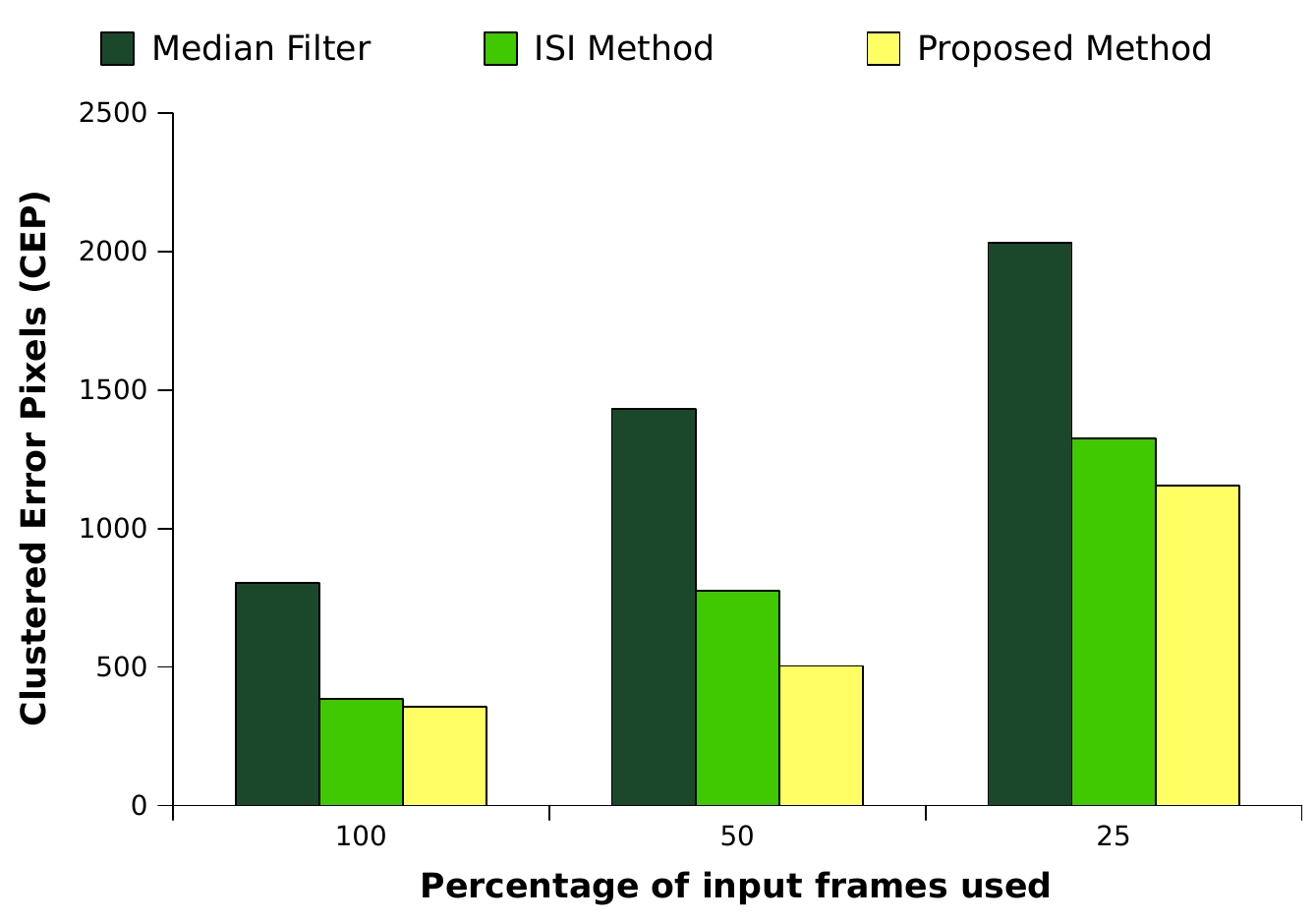}}
    \centerline{\bf (ii)}
  \end{minipage}
   \vspace{0.5ex}
  \caption
    {
    \small
    Averaged values of AGE {\bf (i)} and CEPs {\bf (ii)} obtained by using 100\%, 50\% and 25\% of the input sequences.
    }
  \label{label:age_cep_res}  

\end{center}
\end{figure*}

The Brisbane railway station sequences were not used as their true background was unavailable.
The remaining 10 image sequences were used as listed in Table~\ref{tab:average_Results1}.
To maintain uniformity across sequences, the experiments were conducted using the first 700 frames from each sequence.
The background was estimated in three cases.
In the first case, all  700 frames (100\%) were used to estimate the background.
To evaluate the quality when less frames are available (e.g.~the background needs to be updated more often),
in the second case the sequences were split into halves of 350 frames (50\%) each.
Each sub-sequence was used independently for background estimation and the obtained results were averaged.
In the third case each sub-sequence was further split into halves (i.e.,~25\% of the total length).
Further division of the input resulted in sub-sequences in which parts of the background were always occluded and hence were not utilised.
The averaged AGE and CEP values in all three cases 
are graphically illustrated in Figure~\ref{label:age_cep_res}
and tabulated in Tables~\ref{tab:average_Results1} and~\ref{tab:average_Results2}.
The visual results in Figure~\ref{results_comp123} confirm the objective results,
with the proposed method producing better quality backgrounds than the median filter approach and the ISI method.

\begin{table*}[!tb]

  \begin{center}
    \begin{footnotesize}

      \begin{tabular}{|l|c|c|c|c|c|c|c|c|c|} \hline

     \multirow{2}{*}  & \multicolumn{3}{c|}{\bf{case 1: 100\%}}& \multicolumn{3}{c|}{\bf{case 2: 50\%}} & \multicolumn{3}{c|}{\bf{case 3: 25\%}} 
\\   \cline{2-10}
& \multicolumn{3}{c|}{\bf{ Number of input frames = 700}} & \multicolumn{3}{c|}{\bf{Number of input frames = 350}} &  \multicolumn{3}{c|}{\bf{Number of input frames = 175}}  \\ \cline{2-10}

         {\hspace{25pt}\bf{Sequence}} \multirow{2}{*} &   median  & ISI  & proposed & median  & ISI  & proposed & median  & ISI  & proposed  \\  
      
 & filter & method & method& filter & method & method& filter & method & method \\ \hline 
       
m1.10\_abandoned\_object.avi  &  0.88  & 0.88  & 0.42  & 1.45  & 1.08  & 0.70  & 1.27  & 1.3   & 1.25 \\ \hline

m1.16\_abandoned\_object.avi  & 2.02   & 1.69  & 1.93  & 2.06  & 2.03  & 2.25  & 2.38  & 2.36  & 2.65 \\ \hline

m1.15\_abandoned\_object.avi  &  0.50  & 0.59  & 1.03  & 0.51  & 0.64  & 0.79  & 1.26  & 1.1   & 0.87 \\ \hline

OneStopEnter1cor.mpg          & 0.99   & 0.98  & 0.85  & 0.50  & 0.39  & 0.59  & 0.65  & 0.63  & 0.73 \\ \hline

OneStopEnter2cor.mpg          & 1.37   & 1.16  & 0.82  & 1.04  & 0.91  & 1.06  & 1.23  & 1.06  & 1.13 \\ \hline

OneStopNoEnter1cor.mpg        &  0.90  & 0.96  & 0.21  & 0.56  & 0.92  & 0.42  & 1.65  & 1.44  & 0.49  \\ \hline        

OneStopNoEnter2cor.mpg        & 1.01   & 1.62  & 0.53  & 2.44  & 1.67  & 1.40  & 2.99  & 2.15  & 1.92   \\ \hline    

OneStopMoveEnter1cor.mpg      & 3.69   & 2.15  & 0.73  & 6.37  & 2.45  & 1.53  & 7.31  & 4.02  & 4.92   \\ \hline

OneStopMoveNoEnter2cor.mpg    & 0.64   & 0.49  & 0.81  & 0.94  & 1.01  & 0.79  & 1.87  & 1.45  & 1.19 \\ \hline        

TwoEnterShop1cor.mpg          & 2.12   & 1.86  & 1.85  & 3.49  & 3.21  & 1.51  & 4.35  & 4.66  & 3.38   \\ \hline    

{\hspace{25pt}\bf{Average}}   &\bf{1.41}  &\bf{1.24}  &   \bf{0.92} & \bf{1.87} & 
\bf{1.61} & \bf{1.10} &  \bf{2.7}   & \bf{2.37} & \bf{1.85}

\\ \hline
        
      \end{tabular}
    \end{footnotesize}
    \vspace{1ex}
    \caption
     {
     \small
     Averaged grey-level error (AGE) results from experiments on 10 image sequences.
     The results under case~2 and case~3
     (using~50\%  and~25\% of the input sequence, respectively)
     were obtained by averaging over the two and four sub-sequences respectively.
     }
    \label{tab:average_Results1}

  \end{center}

\end{table*}

\begin{table*}[!tb]

  \begin{center}
    \begin{footnotesize}

      \begin{tabular}{|l|c|c|c|c|c|c|c|c|c|} \hline
   \multirow{2}{*}  & \multicolumn{3}{c|}{\bf{case 1: 100\%}}& \multicolumn{3}{c|}{\bf{case 2: 50\%}} & \multicolumn{3}{c|}{\bf{case 3: 25\%}} 
\\   \cline{2-10}
& \multicolumn{3}{c|}{\bf{ Number of input frames = 700}} & \multicolumn{3}{c|}{\bf{Number of input frames = 350}} &  \multicolumn{3}{c|}{\bf{Number of input frames = 175}}  \\ \cline{2-10}

         {\hspace{25pt}\bf{Sequence}} \multirow{2}{*} &   median  & ISI  & proposed & median  & ISI  & proposed & median  & ISI  & proposed  \\  
      
 & filter & method & method & filter & method & method & filter & method & method \\ \hline 
       
m1.10\_abandoned\_object.avi  & 258.00  & 208.00  & 0.00    & 976.50  & 423.50   & 133.50  & 664.75  & 673.25  & 660.75   \\ \hline

m1.16\_abandoned\_object.avi  & 455.00  & 320.00  & 322.00  & 463.00  & 333.50   & 467.00  & 358.25  & 378     & 528.75   \\ \hline
 
m1.15\_abandoned\_object.avi  & 0.00    & 95.00   & 86.00   & 0.00    & 92.00    & 38.00   & 773     & 521.75  & 135.25   \\ \hline
    
OneStopEnter1cor.mpg          & 37.00   & 7.00    & 348.00  & 184.50  & 13.00    & 177.00  & 374.5   & 172.5   & 380.50   \\ \hline
    
OneStopEnter2cor.mpg          & 358.00  & 85.00   & 29.00   & 482.00  & 230.50   & 266.00  & 640     & 351.25  & 374.50   \\ \hline

OneStopNoEnter1cor.mpg        &  141.00 & 104.00  & 67.00   & 437.50  & 466.50   & 252.50  & 1224    & 819     & 286.25   \\ \hline        

OneStopNoEnter2cor.mpg        & 103.00  & 406.00  & 35.00   & 1919.50 & 854.00   & 678.00  & 2282.5  & 1224.25 & 1244.00   \\ \hline    
   
OneStopMoveEnter1cor.mpg      & 3931.00 & 1196.00 & 714.00  & 5756.00 & 2503.00  & 1289.50 & 8365.25 & 4622.25 & 3877.75  \\ \hline

OneStopMoveNoEnter2cor.mpg    & 257.00  & 63.00   & 232.00  & 574.50  & 348.50   & 259.00  & 1169.25 & 697.75  & 654.50      \\ \hline        

TwoEnterShop1cor.mpg          & 2487.00 & 1372.00 & 1733.00 & 3534.00 & 2479.50  & 1483.00 & 4468.25 & 3795.5  & 3420.25  \\ \hline    

{\hspace{25pt}\bf{Average}}   &\bf{802.7} &\bf{385.6} &   \bf{356.6}  & \bf{1432.75}  & 
\bf{774.4} &  \bf{504.40}  &  \bf{2031.98}   & \bf{1325.55}  & \bf{1156.25}

\\ \hline

      \end{tabular}
    \end{footnotesize}
    \vspace{1ex}
    \caption
     {
     \small
     As per Table~\ref{tab:average_Results1}, but using clustered error pixels (CEPs) as the error measure.
     }
    \label{tab:average_Results2}

  \end{center}

\end{table*}

\clearpage
\clearpage

\subsubsection{Sensitivity of {$\mathcal{T}_1$}}
\label{subsec:T1Sensitivity}

\begin{figure}[!tb]
\begin{center}
  \begin{minipage}{0.4\textwidth}
    \centerline{\includegraphics[width=1\columnwidth]{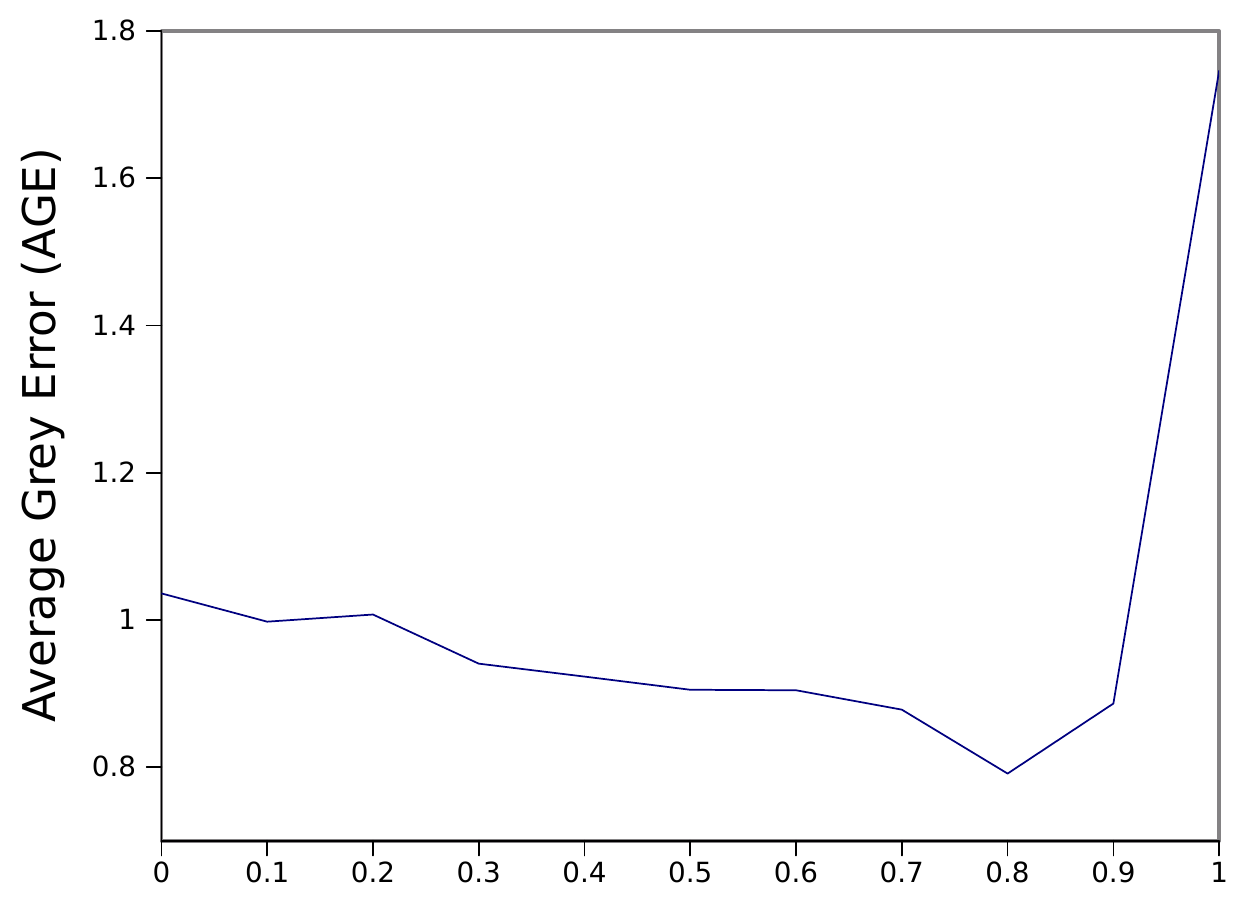}}
  \end{minipage}
   \begin{minipage}{1\columnwidth}
      \centerline{{\bf{$\mathcal{T}_1$}}}
    \end{minipage}
  \caption
    {
    \small
    Effect of {$\mathcal{T}_1$} on AGE, while using a fixed value of {$\mathcal{T}_2$}.
    }
  \label{figure:age_T1_res}  
\end{center}
\end{figure}

To find the optimum value of $\mathcal{T}_1$, we chose a random set of sequences from the CAVIAR dataset,
whose true background was available a-priori and
computed the averaged AGE between the true and estimated 
backgrounds for various values of $\mathcal{T}_1$ as indicated in Figure~\ref{figure:age_T1_res}.
As shown, the optimum value (minimum error) was obtained at $\mathcal{T}_1$ = 0.8.

\subsection{Evaluation by Foreground Segmentation}
\label{sec_Evaluation_by_Foreground_Segmentation}

In order to show the proposed method aids in better segmentation results, we objectively
evaluated the performance of a segmentation algorithm (via background subtraction) 
on the Wallflower dataset.
We note that the proposed method is primarily designed to deal with static backgrounds,
while Wallflower contains both static and dynamic backgrounds.
As such, Wallflower might not not optimal for evaluating the efficacy of the proposed algorithm in its intended domain,
however it can nevertheless be used to provide some suggestive results as to the performance in various conditions.

For foreground object segmentation estimation, we use a Gaussian based background subtraction method
where each background pixel is modeled using a Gaussian distribution.
The parameters of each Gaussian (i.e., the mean and variance) are initialised
either directly from a training sequence, or via the proposed MRF-based background estimation method
(i.e. using labels yielding the maximum value of the posterior probability
described in Eqn.~(\ref{eq:bayestheoremlog}) and their corresponding variances, respectively). 
The median filter and ISI~\cite{wang2006nrs} methods were not used
since they do not define how to compute pixel variances of their estimated background. 

For measurement of foreground segmentation accuracy, 
we use the similarity measure adopted by Maddalena and Petrosino~\cite{maddalena2008self},
which quantifies how similar the obtained foreground mask is to the ground-truth.
The measure is defined as:

\begin{equation}
  \mbox{\it similarity} = \frac{tp}{tp+fp+fn}
  \label{similarity_measure}
\end{equation}%

\noindent
where {\it similarity}~{\small $\in [0,1]$},
while {\small $tp$}, {\small $fp$} and {\small $fn$}
are total number of true positives, false positives and false negatives (in terms of pixels), respectively.
The higher the {${\it similarity}$} value,
the better the segmentation result.
We note that the {\it similiarity} measure is related to precision and recall metrics~\cite{davis2006relationship}.

The parameter settings were the same as used for measuring the standalone performance (Section~\ref{sec_Standalone Evaluation}).
The relative improvements in {\it similarity} resulting from the use of the MRF-based parameter estimation
in comparison to direct parameter estimation are listed in Table~\ref{tab:similarity value}.

We note that each of the Wallflower sequences addresses one specific problem,
such as dynamic background, sudden and gradual illumination variations, camouflage, and bootstrapping.
As mentioned earlier, the proposed method is primarily designed for static background estimation (bootstrapping).
On the `Bootstrap' sequence, characterised by severe background occlusion we register a significant improvement of over 62\%.
On the other sequences, the results are only suggestive and need not always yield high {\it similarity} values.
For example, we note a degradation in the performance on `TimeOfDay' sequence. 
In this sequence, there is steady increase in the lighting intensity from dark to bright, due to which
identical labels were falsely treated as `unique'. As a result, estimated background labels variance appeared
to be smaller than the true variance of the background, which in turn resulted in surplus false positives.
Overall, MRF based background initialisation over 6 sequences achieved an average percentage improvement 
in {\it similarity} value of 16.67\%.

\begin{table}[!tb]
  \begin{center}
  \begin{small}
    \begin{tabular}{|c||c|c|c|} \hline
      {\bf Wallflower}                &  {\bf Relative improvement} \\ 
      {\bf Sequence}                  &  {\bf in {${\it similarity}$} (Eqn.~\ref{similarity_measure})}   \\ \hline
                       
      { WavingTrees}                  & ~34\%         \\ \hline
      {  ForegroundAperture}          & ~~6\%         \\ \hline
      {  LightSwitch}                 & ~~1\%         \\ \hline
      {  Camouflage}                  & ~20\%         \\ \hline
      {  Bootstrap}                   & ~62\%         \\ \hline
      {  TimeOfDay}                   & -23\%         \\ \hline
      { \bf Average}                  & {\bf 16.67}\% \\ \hline      
      
    \end{tabular}
  \end{small}
  \vspace{1ex}
  \caption
    {
    \small
    Relative percentage improvement in foreground segmentation {${\it similarity}$} (Eqn.~\ref{similarity_measure}),
    obtained on the Wallflower dataset,
    resulting from the use of the MRF-based parameter estimation
    in comparison to direct parameter estimation.
    The similarity value of \textit{moved object} sequence 
    turns out to be zero (due to the absence of true positives in its ground-truth) and is therefore not listed.
    }
  \label{tab:similarity value}
  \end{center}
\end{table}

\subsection{Additional Observations}
\label{subsec_Additional_Observations}

We noticed (via subjective observations) that all background estimation algorithms perform reasonably well
when foreground objects are always in motion
(i.e., in cases where the background is visible for a longer duration when compared to the foreground).
In such circumstances, a median filter is perhaps sufficient to reliably estimate the background.
However, accurate estimation by the median filter and the ISI method becomes problematic if the above condition is not satisfied.
This is the main area where the proposed algorithm is able to estimate the background with considerably better quality.

The proposed algorithm sometimes mis-estimates the background in cases
where the true background is characterised by strong edges while the occluding foreground object is smooth (uniform intensity value) 
and has intensity value similar to that of the background (i.e., low contrast between the foreground and the background).
Under these conditions, the energy potential of the label containing the foreground object is smaller
(i.e., smoother spectral response)  than that of the label corresponding to the true background.

From our experiments we found the memory footprint to store the state space of all the nodes
is on average only 5\% of the memory required for storing all the frames.
This is in contrast to existing algorithms,
which typically require the storage of all the frames before processing can begin.

We conducted additionally experiments on image sequences represented in other colour spaces, such as RGB and YUV,
and evaluated the overall posterior as the sum of individual posteriors evaluated on each channel independently. 
The results were marginally better than those obtained using greyscale input.
We conjecture that this is because the spatial continuity of structures within a scene are well represented in greyscale.

\section{Main Findings and Future Work}
\label{sec:Conclusion}

In this paper we proposed a background estimation algorithm in an MRF framework 
that is able to accurately estimate the static background
from cluttered surveillance videos containing image noise as well as foreground objects.
The objects may not always be in motion or may occlude the background for much of the time.

The contributions include the way we define the neighbourhood system,
the cliques and the formulation of clique potential which characterises
the spatial continuity by analysing data in the spectral domain. 
Furthermore, the proposed algorithm has several advantages,
such as computational efficiency and low memory requirements due to sequential processing of frames.
This makes the algorithm possibly suitable for implementation on embedded systems,
such as smart cameras~\cite{mustafah2007smart,wolf2002sce}.

The performance of the algorithm is invariant to moderate illumination changes,
as we consider only AC coefficients of the DCT in the computation of the energy potential
defined by Eqn.~(\ref{eq:energypotential}).
However, the similarity criteria defined by Eqns.~(\ref{corr_ptrk})
and~(\ref{diff_ptrk}) creates multiple representatives for the same visually identical block.
Tackling this problem efficiently is part of further research. We also intend
to extend this work to estimate background models of non-static backgrounds.

Experiments on real-life surveillance videos indicate that
the algorithm obtains considerably better background estimates
(both objectively and subjectively) than methods based 
on median filtering and finding intervals of stable intensity.
Furthermore, segmentation of foreground objects on the Wallflower dataset was 
also improved when the proposed method was used to initialise
the background model based on a single Gaussian.
We note that the proposed background estimation algorithm can be combined
with almost any foreground segmentation technique,
such as \cite{matsuyama2006background,reddy2010adaptive}.

\newpage
\balance

\section*{Acknowledgements}

The authors thank Prof.~Terry Caelli for useful discussions and suggestions.
NICTA is funded by the Australian Government
via the Department of Broadband, Communications and the Digital Economy,
as well as the Australian Research Council through the ICT Centre of Excellence program.

\bibliographystyle{IEEEtran}
\bibliography{references}

\end{document}